\documentclass[review]{elsarticle}

\usepackage{url}            
\usepackage{booktabs}       
\usepackage{amsfonts}       
\usepackage{nicefrac}       
\usepackage{microtype}      
\usepackage{multirow}
\usepackage{xcolor}
\usepackage{graphicx}
\usepackage[position=b]{subcaption}
\usepackage{amsmath}
\usepackage{nccmath}
\usepackage[normalem]{ulem}
\usepackage[ruled,vlined]{algorithm2e}
\usepackage{arydshln}  
\usepackage{natbib}

\usepackage{silence}
\journal{Neural Networks (Accepted)}

\begin{document}

\begin{frontmatter}

\title{Stacked DeBERT: All Attention in Incomplete Data for Text Classification} 

\author{Gwenaelle Cunha Sergio$^{1}$ and Minho Lee$^{2}$}
\address{$^{1}$ School of Electronics and Electrical Engineering \\
         $^{2}$ Department of Artificial Intelligence \\ 
         Kyungpook National University, Daegu, 41566, South Korea}




\begin{abstract}
In this paper, we propose \textbf{Stacked DeBERT}, short for \textbf{Stacked} \textbf{De}noising \textbf{B}idirectional \textbf{E}ncoder \textbf{R}epresentations from \textbf{T}ransformers. This novel model improves robustness in incomplete data, when compared to existing systems, by designing a novel encoding scheme in BERT, a powerful language representation model solely based on attention mechanisms. Incomplete data in natural language processing refer to text with missing or incorrect words, and its presence can hinder the performance of current models that were not implemented to withstand such noises, but must still perform well even under duress. This is due to the fact that current approaches are built for and trained with clean and complete data, and thus are not able to extract features that can adequately represent incomplete data. Our proposed approach consists of obtaining intermediate input representations by applying an embedding layer to the input tokens followed by vanilla transformers. These intermediate features are given as input to novel denoising transformers which are responsible for obtaining richer input representations. The proposed approach takes advantage of stacks of multilayer perceptrons for the reconstruction of missing words' embeddings by extracting more abstract and meaningful hidden feature vectors, and bidirectional transformers for improved embedding representation. We consider two datasets for training and evaluation: the Chatbot Natural Language Understanding Evaluation Corpus and Kaggle's Twitter Sentiment Corpus. Our model shows improved F1-scores and better robustness in informal/incorrect texts present in tweets and in texts with Speech-to-Text error in the sentiment and intent classification tasks.~\footnote{\url{https://github.com/gcunhase/StackedDeBERT}}
\end{abstract}


\end{frontmatter}


\section{Introduction}
\label{sec:intro}
Understanding a user's intent and sentiment is of utmost importance for current intelligent chatbots to respond appropriately to human requests. However, current systems are not able to perform to their best capacity when presented with incomplete data, meaning sentences with missing or incorrect words. This scenario is likely to happen when one considers human error done in writing. In fact, it is rather naive to assume that users will always type fully grammatically correct sentences~\cite{panko2008thinking}. This has been aggravated with the advent of the internet and social networks, which allowed language and modern communication to be been rapidly transformed~\cite{al2018evolution,grieve2018mapping}. Take Twitter for instance, where information is expected to be readily communicated in short and concise sentences with little to no regard to correct sentence grammar or word spelling~\cite{sirucek2010twitter}.

Further motivation can be found in Automatic Speech Recognition (ASR) applications, where high error rates prevail and pose an enormous hurdle in the broad adoption of speech technology by users worldwide~\cite{errattahi2018automatic}. This is an important issue to tackle because, in addition to more widespread user adoption, improving Speech-to-Text (STT) accuracy diminishes error propagation to modules using the recognized text. With that in mind, in order for current systems to improve the quality of their services, there is a need for development of robust intelligent systems that are able to understand a user even when faced with incomplete representation in language. 

The advancement of deep neural networks have immensely aided in the development of the Natural Language Processing (NLP) domain. Tasks such as text generation, sentence correction, image captioning and text classification, have been possible via models such as Convolutional Neural Networks and Recurrent Neural Networks~\cite{sergio2018temporal,vinyals2015show,kim2014convolutional}. More recently, state-of-the-art results have been achieved with attention models, more specifically Transformers~\cite{vaswani2017attention}. Current approaches for Text Classification tasks focus on efficient embedding representations. Kim et al.~\cite{kim2016intent} use semantically enriched word embeddings to make synonym and antonym word vectors respectively more and less similar in order to improve intent classification performance. Devlin et al.~\cite{devlin2018bert} propose Bidirectional Encoder Representations from Transformers (BERT), a powerful bidirectional language representation model based on Transformers, achieving state-of-the-art results on eleven NLP tasks~\cite{wang2018glue}, including sentiment text classification. Concurrently, Shridhar et al.~\cite{shridhar2018subword} also reach state of the art in the intent recognition task using Semantic Hashing for feature representation followed by a neural classifier. All aforementioned approaches are, however, applied to datasets based solely on complete data.

Incompleteness in data can refer to noise in either the input~\cite{vateekul2016study,tiwari2017normalization,joshi2018twitter,lourentzou2019adapting,vinciarelli2005noisy,agarwal2007much,shrestha2019using} or in the labels~\cite{nigam2000text,ramakrishnan2005model,tsuboi2008training}. In this work, we focus on the former. More specifically, we focus on noisy text~\cite{subramaniam2010noisy} obtained from social networks~\cite{vateekul2016study,tiwari2017normalization,joshi2018twitter,lourentzou2019adapting} and through ASR processing techniques~\cite{vinciarelli2005noisy,agarwal2007much,shrestha2019using}. 

Studies on social media text classification mainly focus on normalizing the data into a clean or standard form before classification for improved performance, instead of considering the data incompleteness as it is. Vateekul and Koomsubha~\cite{vateekul2016study} apply pre-processing techniques on Thai Twitter data and evaluate sentiment classification performance on two Deep Learning models: Long Short-Term Memory and Dynamic Convolutional Neural Network. Joshi and Deshpande~\cite{joshi2018twitter} also apply extensive pre-processing steps to noisy social media text and extract their n-gram features in order to evaluate their sentiment on probabilistic classifiers, namely Naive Bayes and Maximum Entropy. More recently, researchers have proposed using encoder-decoder frameworks to perform social media text normalization. Tiwari and Naskar~\cite{tiwari2017normalization} achieve good results by training their model with synthetic data and later using transfer learning on the WNUT 2015 shared task dataset. Lourentzou et al.~\cite{lourentzou2019adapting} also achieve comparable results by proposing a hybrid word-character attention-based encoder-decoder model. Some researchers have additionally attempted at improving noisy text embeddings. Barbosa and Feng~\cite{barbosa2010robust} propose a two-step model for sentiment classification in tweets. The authors first classify the tweets into subjective or objective and then further classify the subjective tweets as positive or negative. They obtain a feature vector from tweets using additional information, namely meta-information from words and written characteristics from tweets, and compare it with other raw word representations such as n-grams. The authors then show improved classification results using Support Vector Machine (SVM) as a classifier.

Other researchers, such as Vinciarelli~\cite{vinciarelli2005noisy} and Agarwal et al.~\cite{agarwal2007much}, focus on the simulation of ASR noise in text and the effect that their introduction has in the classification performance of Naive Bayes and SVM. Shrestha et al.~\cite{shrestha2019using} expand that research by investigating the decay in performance in logistic regression and neural networks. These works focus on investigating the impact in the performance of an existing model given incomplete data. More recently, with the introduction of the transformer~\cite{vaswani2017attention} based model BERT~\cite{devlin2018bert}, researchers~\cite{hrinchuk2020correction,liao2020improving} have changed the focus to improving a model's performance by extracting more robust  input representation. Hrinchuk et al.~\cite{hrinchuk2020correction} and Liao et al.~\cite{liao2020improving} propose to use a transformer encoder-decoder architecture for the task of ASR correction.

Realizing the need for further research in the area of noisy text classification, we make it the focus of this paper. In this task, the model aims to identify the user's intent or sentiment by analyzing a sentence with missing and/or incorrect words. In the sentiment classification task, the model aims to identify the user's sentiment given a tweet, written in informal language and without regards for sentence correctness. As for the incomplete data problem, we approach it as a reconstruction or imputation task~\cite{pratama2016review}. Vincent et al.~\cite{vincent2008extracting,vincent2010stacked} and Glorot et al.~\cite{glorot2011deep} propose to reconstruct clean data from their noisy version by mapping the input to meaningful representations. This approach has also been shown to outperform other models, such as predictive mean matching, random forest, SVM and Multiple imputation by Chained Equations (MICE), at missing data imputation tasks~\cite{gondara2017multiple,costa2018missing}.

Researchers in the noisy text and missing data imputation areas have shown that meaningful feature representation of data is of utter importance for high performance achieving methods. We propose a model that combines the power of BERT in the NLP domain and the strength of denoising strategies in incomplete data reconstruction to tackle the tasks of incomplete intent and sentiment classification. This enables the implementation of a novel encoding scheme, more robust to incomplete data, called Stacked Denoising BERT or Stacked DeBERT. Our approach consists of obtaining richer input representations from input tokens by stacking denoising transformers on an embedding layer with vanilla transformers. The embedding layer and vanilla transformers extract intermediate input features from the input tokens, and the denoising transformers are responsible for obtaining richer input representations from them. By improving BERT with stronger denoising abilities, we are able to reconstruct missing and incorrect words' embeddings and improve classification accuracy. To summarize, our contribution is three-fold:
\begin{itemize}
    \item Implementation of a novel encoding scheme to obtain richer embedding representations. This is done by reconstructing hidden embeddings from noisy input and using information from both incomplete and complete data during training.
    \item Improvement on the robustness and performance of BERT when applied to incomplete data, in particular noisy user-generated texts obtained from Twitter and noisy ASR text.
    \item Open-source code and release of corpora used in the tasks of incomplete intent and sentiment classification from incorrect sentences. 
\end{itemize}

The remainder of this paper is organized in four sections, with Section~\ref{sec:proposed_model} explaining the proposed model. This is followed by Section~\ref{sec:dataset} which includes a detailed description of the dataset used for training and evaluation purposes and how it was obtained. Section~\ref{sec:experiments} covers the baseline models used for comparison, training specifications and experimental results and Section~\ref{sec:discussion} has further analysis and discussion on the results. Finally, Section~\ref{sec:conclusion} wraps up this paper with conclusion and future works.

\section{Proposed model}
\label{sec:proposed_model}

\begin{figure}[ht!]
\centering
\begin{subfigure}[t!]{.85\textwidth}
\includegraphics[width=\textwidth]{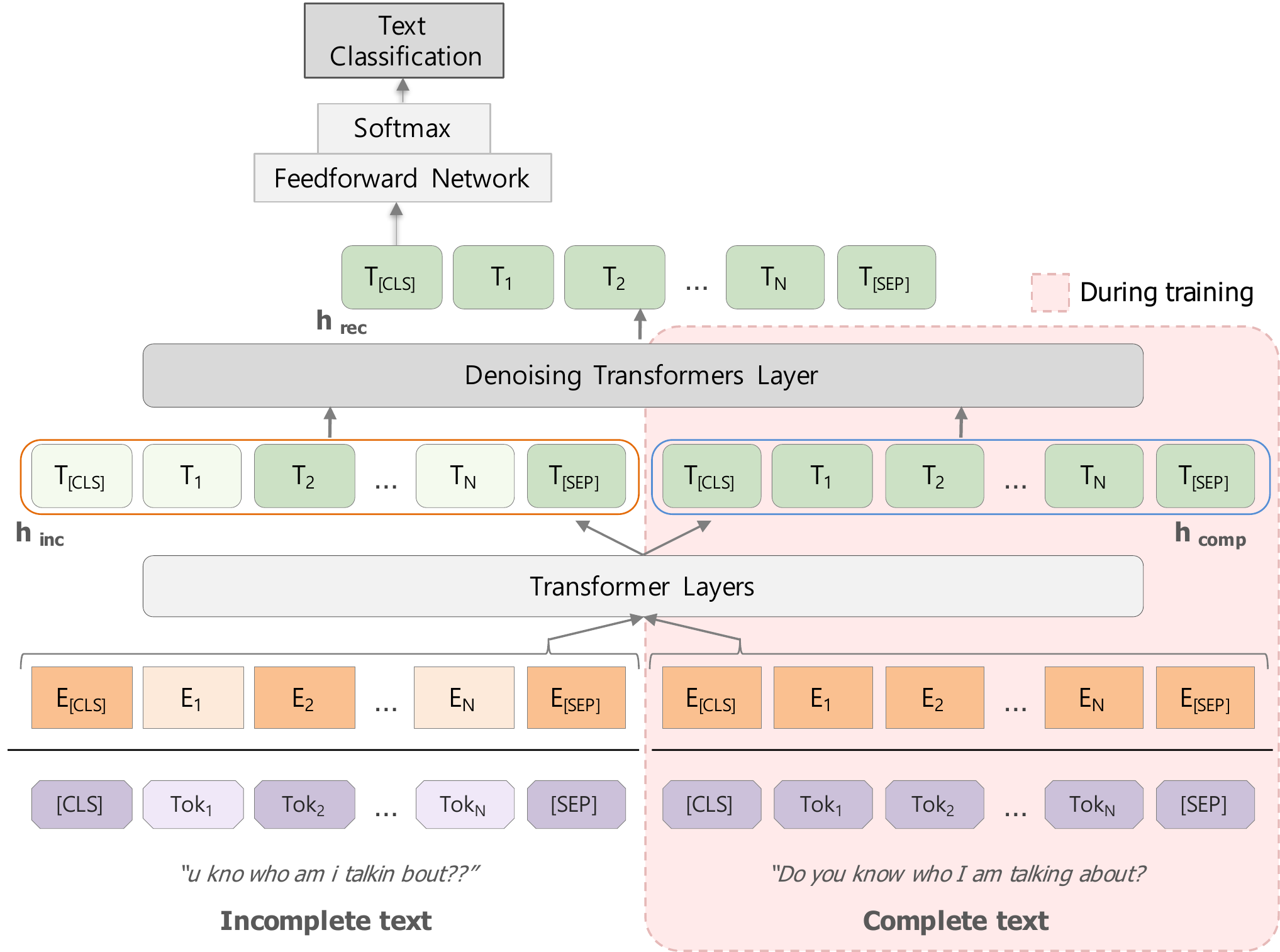}
\caption{Model architecture}\label{fig:proposed_model_architecture}
\end{subfigure}
\begin{subfigure}[t!]{.41\textwidth}
\includegraphics[width=0.85\textwidth]{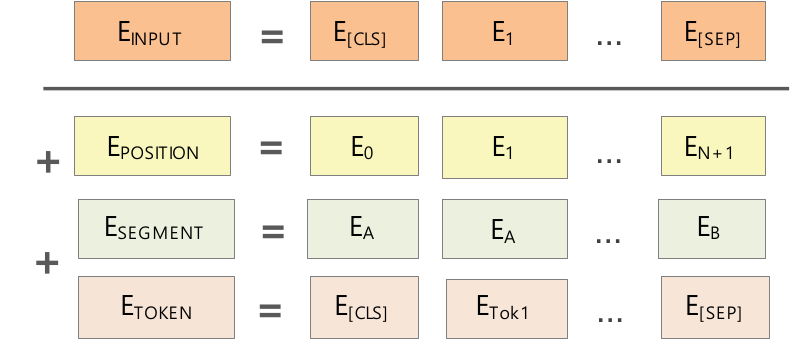}
\caption{Embedding layers}\label{fig:embedding_layers}
\end{subfigure}
~
\begin{subfigure}[t!]{.44\textwidth}
\includegraphics[width=\textwidth]{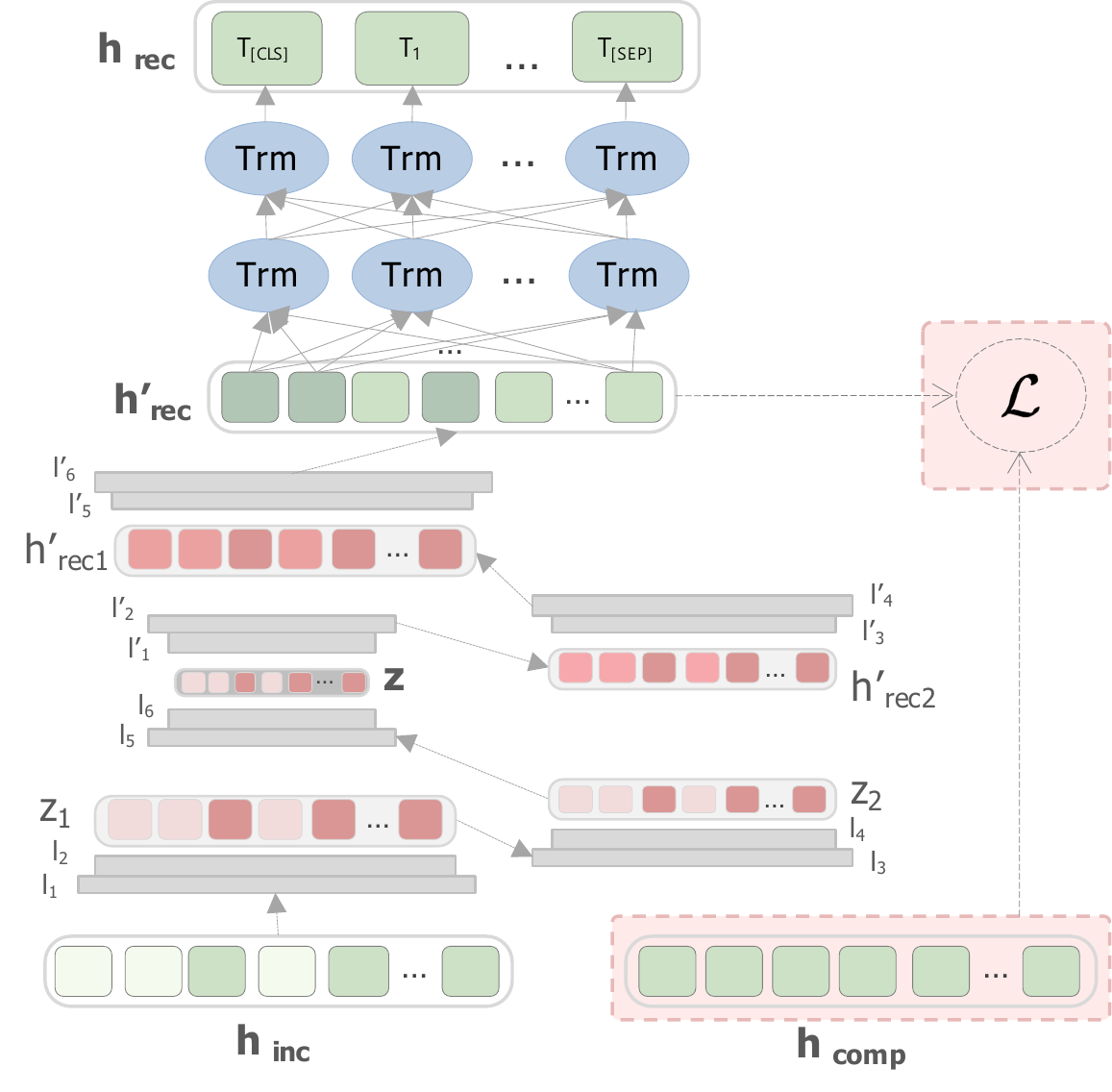}
\caption{Denoising transformers layer}\label{fig:denoising_transformers}
\end{subfigure}
\caption{The proposed model Stacked DeBERT (a) is organized in three modules: embedding (b), conventional bidirectional transformers, and denoising bidirectional transformers (c). During training, our model obtains the intermediate embeddings for both incomplete and complete data, represented by $h_{inc}$ and $h_{comp}$ respectively, with conventional BERT and trains stacks of multilayer perceptrons to partially reconstruct the intermediate embedding $h_{inc}$ into $h'_{rec}$. This embedding is then given to bidirectional transformers to generate the final reconstructed embedding $h_{rec}$. In the end, a more robust \texttt{[CLS]} token is obtained for text classification. During testing, only the incomplete text is used.} \label{fig:proposed_model}
\end{figure}

We propose Stacked Denoising BERT (DeBERT) as a novel encoding scheming for the task of incomplete intent classification and sentiment classification from incorrect sentences, such as tweets and text with STT error. The proposed model, illustrated in Fig. \ref{fig:proposed_model}, is structured as a stacking of embedding layers and vanilla transformer layers, similarly to the conventional BERT~\citep{devlin2018bert}, followed by layers of novel denoising transformers. It is important to note that part of the reason why BERT is such a powerful language model is due to its pre-training objectives: next sentence prediction (NSP) and masked language model (MLM). The first is used to help BERT understand sentence continuity by forcing it to predict whether two given sentences are in the correct order. This is very important to model multiple-sentences inputs, but in our case, we only deal with single sentences inputs. The second pre-training objective, on the other hand, forces the model to predict random masked words, meaning that it indirectly models data incompleteness to a certain degree. However, BERT does not directly handle incorrect words in the input, resulting in a reduction in performance when faced with noisy user-generated text such as Twitter or text obtained through ASR processing techniques. We aim to ameliorate that limitation and improve the robustness and efficiency of BERT when applied to incomplete data by reconstructing hidden embeddings from sentences with missing words. By reconstructing these hidden embeddings, we are able to improve the encoding scheme in BERT.

The initial part of the model is the conventional BERT, a multi-layer bidirectional Transformer encoder and a powerful language model. During training, BERT is fine-tuned on the incomplete text classification corpus (see Section \ref{sec:dataset}). The first layer pre-processes the input sentence by making it lower-case and by tokenizing it. It also prefixes the sequence of tokens with a special character \texttt{[CLS]} and sufixes each sentence with a \texttt{[SEP]} character. It is followed by an embedding layer used for input representation, with the final input embedding being a sum of token embedddings, segmentation embeddings and position embeddings. The first one, token embedding layer, uses a vocabulary dictionary to convert each token into a more representative embedding. The segmentation embedding layer indicates which tokens constitute a sentence by signaling either 1 or 0. In our case, since our data are formed of single sentences, the segment is 1 until the first \texttt{[SEP]} character appears (indicating segment A) and then it becomes 0 (segment B). The position embedding layer, as the name indicates, adds information related to the token's position in the sentence. This prepares the data to be considered by the layers of vanilla bidirectional transformers, which outputs a hidden embedding that can be used by our novel layers of denoising transformers.

Although BERT has shown to perform better than other baseline models when handling incomplete data, it is still not enough to efficiently handle such data, as mentioned in the beginning of this section. Because of that, there is a need for further improvement of the hidden feature vectors obtained from sentences with missing words. With this purpose in mind, we implement a novel encoding scheme consisting of denoising transformers, which uses information from both incomplete and complete data during training and only incomplete data during testing. During training, the conventional BERT, composed of embedding layers and vanilla transformers (see Fig. \ref{fig:proposed_model}), is used to obtain the intermediate embeddings from both the incomplete and the complete data. These intermediate embeddings, $h_{inc}$ and $h_{comp}$ respectively, are then used to train the denoising transformers layer in a two-step process. The first step is to train stacks of multilayer perceptrons to partially reconstruct the incomplete embedding $h_{inc}$ into the partially, and more meaningful, recovered embedding $h'_{rec}$ through a comparison loss with the complete embedding $h_{comp}$. The second step improves the partial embedding representation by giving $h'_{rec}$ to bidirectional transformers to generate the final reconstructed embedding $h_{rec}$. In the end, a more robust \texttt{[CLS]} token is obtained for text classification.

Note that the intermediate embeddings $h_{inc}$ and $h_{comp}$ both have shape $(N_{bs}, 768, 128)$, where $N_{bs}$ is the batch size, $768$ is the original BERT embedding size for a single token, and $128$ is the maximum sequence length in a sentence. The stacks of multilayer perceptrons are structured as two sets of three layers with two hidden layers each. The first set is responsible for compressing the $h_{inc}$ into a latent-space representation, extracting more abstract features into lower dimension vectors $z_1$, $z_2$ and $\mathbf{z}$ with shape $(N_{bs}, 128, 128)$, $(N_{bs}, 32, 128)$, and $(N_{bs}, 12, 128)$, respectively. This process is shown in Eq. \eqref{eq:autoencoder1}:

\begin{equation} \label{eq:autoencoder1}
\begin{aligned}
&z_1 = W_{l_2} \big(W_{l_1} h_{inc} + b_{l_1}\big) + b_{l_2} = f_{\{W,b\}}\big(h_{inc}\big) \\
&z_2 = W_{l_4} \big(W_{l_3} z_1 + b_{l_3}\big) + b_{l_4} = f_{\{W_{z_1},b_{z_1}\}}\big(z_1\big) \\
&\mathbf{z} = W_{l_6} \big(W_{l_5} z_2 + b_{l_5}\big) + b_{l_6} = f_{\{W_{z_2},b_{z_2}\}}\big(z_2\big) \\
\end{aligned}
\end{equation}
where $f(\cdot)$ is the parameterized function mapping $h_{inc}$ to the hidden state $\mathbf{z}$. The weight matrices and bias are represented by $W$ and $b$ respectively, with the index $l_{i \in \{1\dots6\}}$ indicating its corresponding layer. The second set then respectively reconstructs $z_1$, $z_2$ and $\mathbf{z}$ into $h'_{rec_1}$, $h'_{rec_2}$ and $h'_{rec}$. This process is shown in Eq. \eqref{eq:autoencoder2}:

\begin{equation} \label{eq:autoencoder2}
\begin{aligned}
&h'_{rec_2} = W_{l'_2} \big(W_{l'_1} \mathbf{z} + b_{l'_1}\big) + b_{l'_2} = g_{\{W_{z_2}',b_{z_2}'\}}\big(\mathbf{z}\big)\\
&h'_{rec_1} = W_{l'_4} \big(W_{l'_3} h'_{rec_2} + b_{l'_3}\big) + b_{l'_4} = g_{\{W_{z_1}',b_{z_1}'\}}\big(h'_{rec_2}\big)\\
&h'_{rec} = W_{l'_6} \big(W_{l'_5} h'_{rec_1} + b_{l'_5}\big) + b_{l'_6} = g_{\{W',b'\}}\big(h'_{rec_1}\big)\\
\end{aligned}
\end{equation}
where $g(\cdot)$ is the parameterized function that reconstructs $\mathbf{z}$ as $h'_{rec}$. In other words, $h'_{rec}$ is the intermediate reconstructed embeddings, obtained by reconstructing the latent-space representation $\mathbf{z}$ through a set of stacks of multilayer perceptrons. The weight matrices and bias are represented by $W$ and $b$ respectively, with the index $l'_{i \in \{1\dots6\}}$ indicating its corresponding layer.

The intermediate reconstructed embedding $h'_{rec}$ is compared with the complete embedding $h_{comp}$ through a mean square error loss function, as shown in Eq. \eqref{eq:loss}:

\begin{equation} \label{eq:loss}
\begin{aligned}
\mathcal{L}\big(h'_{rec}, h_{comp}\big) &= \frac{1}{N_{bs}} \sum_{i=1}^{N_{bs}} \Big(h'_{rec}(i) - h_{comp}(i)\Big)^2 \\
\end{aligned}
\end{equation}

After reconstructing the correct hidden embeddings from the incomplete sentences, the correct hidden embeddings are given to bidirectional transformers to generate the final reconstructed input representations $h_{rec}$. The model is then fine-tuned in an end-to-end manner on the incomplete text classification corpus.

Classification is done with a feedforward network and softmax activation function. Softmax $\sigma$ is a discrete probability distribution function for $N_C$ classes, with the sum of the classes probability being 1 and the maximum value being the predicted class. The predicted class can be mathematically calculated as in Eq. \eqref{eq:intent_class}:
\begin{equation} \label{eq:intent_class}
\begin{aligned}
\hat{y} = \underset{i \in 1... N_C}{\arg\max} \ \ p_i = \underset{i \in 1... N_C}{\arg\max} \ \ \sigma(o_i) = \underset{i \in 1... N_C}{\arg\max} \ \ \frac{e^{o_i}}{\sum_{k=1}^{N_C} e^{o_i}} \\
\end{aligned}
\end{equation}
where $o = W t + b$, the output of the feedforward layer used for classification.

\section{Dataset}
\label{sec:dataset}

\subsection{Twitter Sentiment Classification}
In order to evaluate the performance of our model, we need access to a naturally noisy dataset with real human errors. Poor quality texts obtained from Twitter, called tweets, are then ideal for our task. For this reason, we choose Kaggle's two-class Sentiment140 dataset~\cite{go2009twitterSentiment140}\footnote{\url{https://www.kaggle.com/kazanova/sentiment140}}, which consists of spoken text being used in writing and without strong consideration for grammar or sentence correctness. Thus, it has many mistakes, as specified in Table \ref{tbl:twitter_mistakes}~\cite{lourentzou2019adapting}.

\begin{table}[ht!]
  \caption{Types of mistakes on the Twitter dataset.}
  \label{tbl:twitter_mistakes}
  \bigskip
  \centering
  \scalebox{0.85}{
  \begin{tabular}{ll} 
    \toprule
    \textbf{Mistake type} & \textbf{Examples} \\
    \midrule
    Spelling & \textit{``teh"} (the), \textit{``correclty"} (correctly), \textit{``teusday"} (Tuesday) \\
    Casual pronunciation & \textit{``wanna"} (want to), \textit{``dunno"} (don't know) \\
    Abbreviation & \textit{``Lit"} (Literature), \textit{``pls"} (please), \textit{``u"} (you), \textit{``idk"} (I don't know) \\
    Repeteated letters & \textit{``thursdayyyyyy"}, \textit{``sleeeeeeeeeep"} \\
    Onomatopoeia & \textit{``Woohoo"}, \textit{``hmmm"}, \textit{``yaay"} \\
    Others & \textit{``im"} (I'm), \textit{``your/ur"} (you're), \textit{``ryt"} (right)\\
    \bottomrule
  \end{tabular}}
\end{table}

Even though this corpus has incorrect sentences and their emotional labels, they lack their respective corrected sentences, necessary for the training of our model. In order to obtain this missing information, we use Amazon Mechanical Turk (MTurk)~\cite{buhrmester2011amazon}, a paid marketplace for Human Intelligence Tasks (HITs) which allows for anonymity between ``requesters” and ``workers”. This ensures that the requester is not able to influence the worker into answering a survey the way they want, reducing bias and allowing for believable results to be obtained. In this work, we simply use MTurk to outsource native English speakers to obtain the correct sentences from the original tweets. More specifically, human annotators are given a list of original tweets and they are asked to correct them with as little change as possible and without inserting any extra punctuation marks unless absolutely necessary, following guidelines for noisy text normalization~\cite{lourentzou2019adapting}. After getting the data back, we manually check if the corrections are acceptable, otherwise we post another HIT. We claim this is unbiased because this is only dependent on the English language, not on the sentiment or corrector's background. Some examples are shown in Table \ref{tbl:twitter_examples}.

\begin{table}[ht!]
  \caption{Examples of original tweets and their corrected version.}
  \label{tbl:twitter_examples}
  \bigskip
  \centering
  \scalebox{0.85}{
  \begin{tabular}{lp{6.5cm}} 
    \toprule
    \textbf{Original tweet} & \textbf{Corrected tweet} \\
    \midrule
    ``goonite  sweet dreamz" & ``Good night, sweet dreams."\\
    ``well i dunno..i didnt give him an ans yet" & ``Well I don't know, I didn't give him an answer yet." \\
    ``u kno who am i talkin bout??" & ``Do you know who I am talking about?" \\
    \bottomrule
  \end{tabular}}
\end{table}

After obtaining the correct sentences, our two-class dataset~\footnote{Available at \url{https://github.com/gcunhase/StackedDeBERT}} has class distribution as shown in Table \ref{tbl:corpus_info_twitter}. There are 200 sentences used in the training stage, with 100 belonging to the \textit{positive} sentiment class and 100 to the \textit{negative} class, and 50 samples being used in the evaluation stage, with 25 negative and 25 positive. This totals in 250 samples, with incorrect and correct sentences combined. Since our goal is to evaluate the model's performance and robustness in the presence of noise, we only consider incorrect data in the testing phase. Note that BERT is a pre-trained model, meaning that small amounts of data are enough for appropriate fine-tuning.

\begin{table}[ht!]
 \caption{Details about our Twitter Sentiment Classification dataset, composed of incorrect and correct data.}
 \label{tbl:corpus_info_twitter}
 \bigskip
 \centering
 \scalebox{0.95}{
 \begin{tabular}{lllll}
    \toprule
    Dataset & Sentiment & Train & Test & Total \\
    \midrule
    \multirow{3}{*}{Sentiment140} & Negative (0) & 100 & 25 & 125 \\
    & Positive (1) & 100 & 25 & 125 \\
    \cmidrule(r){2-5}
    & Total & 200 & 50 & 250 \\
    \bottomrule
 \end{tabular}}
\end{table}

\subsection{Intent Classification from Text with STT Error}

In the intent classification task, we are presented with a corpus that suffers from the opposite problem of the Twitter sentiment classification corpus. In the intent classification corpus, we have the complete sentences and intent labels but lack their corresponding incomplete sentences, and since our task revolves around text classification in incomplete or incorrect data, it is essential that we obtain this information. To remedy this issue, we apply a Text-to-Speech (TTS) module followed by a Speech-to-Text (STT) module to the complete sentences in order to obtain incomplete sentences with STT error. Due to TTS and STT modules available being imperfect, the resulting sentences have a reasonable level of noise in the form of missing or incorrectly transcribed words. Analysis on this dataset~\footnote{Available at \url{https://github.com/gcunhase/StackedDeBERT}} adds value to our work by enabling evaluation of our model's robustness to different rates of data incompleteness.

The dataset used to evaluate the models' performance is the Chatbot Natural Language Understanding (NLU) Evaluation Corpus, introduced by Braun et al.~\cite{braun2017evaluating} to test NLU services. It is a publicly available~\footnote{\url{https://github.com/sebischair/NLU-Evaluation-Corpora}} benchmark and is composed of sentences obtained from a German Telegram chatbot used to answer questions about public transport connections. The dataset has two intents, namely \textit{Departure Time} and \textit{Find Connection} with 100 train and 106 test samples, shown in Table \ref{tbl:corpus_info}. Even though English is the main language of the benchmark, this dataset contains a few German station and street names.

\begin{table}[ht!]
  \caption{Details about our Incomplete Intent Classification dataset based on the Chatbot NLU Evaluation Corpus.}
  \label{tbl:corpus_info}
  \bigskip
  \centering
  \scalebox{0.95}{
  \begin{tabular}{lllll}
    \toprule
    Dataset & Intent & Train & Test & Total \\
    \midrule
    \multirow{2}{*}{Chatbot NLU} & Departure Time (0) & 43 & 35 & 98 \\
    & Find Connection (1) & 57 & 71 & 128 \\
    \cmidrule(r){2-5}
    & Total & 100 & 106 & 206 \\
    \bottomrule
  \end{tabular}}
\end{table}

\begin{figure}
    \centering
    \includegraphics[width=0.55\linewidth]{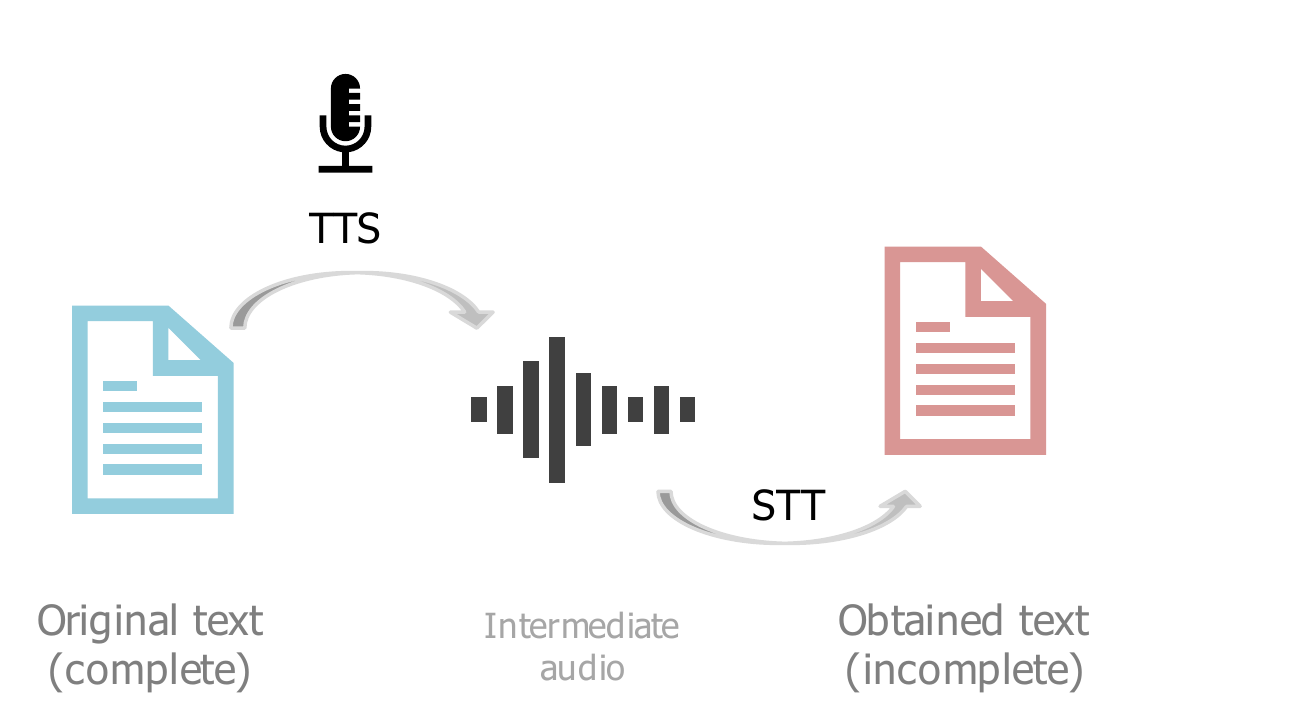}
    \caption{Diagram of 2-step process to obtain dataset with STT error in text.}
    \label{fig:tts_stt_process}
\end{figure}

The incomplete dataset used for training is composed of lower-cased incomplete data obtained by manipulating the original corpora. The incomplete sentences with STT error are obtained in a 2-step process shown in Fig. \ref{fig:tts_stt_process}. The first step is to apply a TTS module to the available complete sentence. Here, we apply \textit{gtts}~\footnote{\url{https://pypi.org/project/gTTS/}}, a Google Text-to-Speech python library, and \textit{macsay}~\footnote{\url{https://ss64.com/osx/say.html}}, a terminal command available in Mac OS as \textit{say}. The second step consists of applying an STT module to the obtained audio files in order to obtain text containing STT errors. The STT module used here was witai~\footnote{\url{https://wit.ai}}, freely available and maintained by Wit.ai. The mentioned TTS and STT modules were chosen according to code availability and whether it's freely available or has high daily usage limitations.

Table \ref{tbl:missing_data} exemplifies a complete and its respective incomplete sentences with different TTS-STT combinations, thus varying rates of missing and incorrect words. The level of noise is denoted by two metrics: inverted BLEU (iBLEU) and Word Error Rate (WER) score. The inverted BLEU score ranges from $0$ to $1$ and is denoted by Eq.~\eqref{eq:ibleu}:

\begin{equation} \label{eq:ibleu}
    iBLEU = 1-BLEU
\end{equation}
where BLEU is a common metric usually used in machine translation tasks~\cite{papineni2002bleu}. We decide to showcase that instead of regular BLEU because it is more indicative to the amount of noise in the incomplete text, where the higher the iBLEU, the higher the noise.

We also provide noise-level measurements in terms of the WER metric, widely used to evaluate automatic speech recognition systems. This metric indicates the amount of effort needed to revert a given sentence into its golden form. In other words, it calculates the number of words deletion $D$, insertion $I$, and substitution $S$, in relation to the total number of words $N$ in the reference, as shown in Eq. \eqref{eq:wer}:
\begin{equation} 
\label{eq:wer}
    WER = \frac{S+D+I}{N}
\end{equation}
Similarly to iBLEU, the lower the WER score, the lower the level of noise in a sentence.

\begin{table}[ht!]
  \caption{Example of sentence from Chatbot NLU Corpus with different TTS-STT combinations and their respective inverted BLEU and WER scores, which denote the level of noise in the text.}
  \label{tbl:missing_data}
  \bigskip
  \centering
  \scalebox{0.75}{
  \begin{tabular}{p{2.5cm}p{6.5cm}p{6.5cm}} 
    \toprule
    \multirow{2}{=}{\textbf{TTS-STT} (iBLEU/WER)} & \multirow{2}{*}{\textbf{Original sentence}} & \multirow{2}{*}{\textbf{With STT error}} \\ \\
    \midrule
    \multirow{5}{=}{gtts-witai (0.44/2.39)} & ``how can i get from garching to milbertshofen?" & ``how can i get from garching to melbourne open."\\
     &  ``how to get from bonner platz to freimann?" & ``how to get from bonner platz to fry."\\
     &  ``prinzregentenplatz to rotkreuzplatz" & ``prince richard replies to recruit plants."\\
     &  ``when does the next u-bahn depart at garching?" & ``when does the next bus depart at garching."\\
    \midrule  
    \multirow{5}{=}{macsay-witai (0.50/3.11)} & ``how can i get from garching to milbertshofen?" & ``how can i get from garching to meal prep."\\
     &  ``how to get from bonner platz to freimann?" & ``how to get from bonner platz to fry."\\
     &  ``prinzregentenplatz to rotkreuzplatz" & ``brandon regional flats to rent."\\
     &  ``when does the next u-bahn depart at garching?" & ``when does the next oakland airport or city."\\
    \bottomrule
  \end{tabular}
  }
\end{table}

\section{Experimental Results}
\label{sec:experiments}

\subsection{Baseline models}
Besides the already mentioned BERT, the following baseline models are also used for comparison.

\paragraph{NLU service platforms} We focus on the three following services, where the first two are commercial services and last one is open source with two separate backends: Google Dialogflow (formerly Api.ai)~\footnote{\url{https://dialogflow.com}}, SAP Conversational AI (formerly Recast.ai)~\footnote{\url{https://cai.tools.sap}} and Rasa (spacy and tensorflow backend)~\footnote{\url{https://rasa.com}}. To the best of our knowledge, these platforms don't have any special measures to handle incomplete data.

\paragraph{Semantic hashing with classifier} Shridhar et al.~\cite{shridhar2018subword} proposed a word embedding method that doesn't suffer from out-of-vocabulary issues. The authors achieve this by using hash tokens in the alphabet instead of a single word, making it vocabulary independent. For classification, classifiers such as Multilayer Perceptron (MLP), Support Vector Machine (SVM) and Random Forest are used. A complete list of classifiers and training specifications are given in Section \ref{sec:training_specs}.

\subsection{Training specifications}
\label{sec:training_specs}
The baseline and proposed models are each trained separately on the incomplete intent classification and Twitter sentiment classification tasks. The intent classification data include three different datasets: complete data and two TTS-STT variants (\textit{gtts-witai} and \textit{macsay-witai}). The sentiment classification data also include three different datasets: original text, corrected text and incorrect with correct texts. The reported F1 scores are the best accuracies obtained from 10 runs using each dataset individually as training data.

\paragraph{NLU service platforms} No settable training configurations available in the online platforms. These frameworks simply take a given training dataset, without distinction between clean or incorrect data, and train and test it on their own platform. The only difference is in the data format it accepts, which changes depending on the platform or code. For example, Rasa uses json format, whereas Dialogflow and SAP uses csv format.

\paragraph{Semantic hashing with classifier} Trained on 3-gram, feature vector size of 768 as to match the BERT embedding size, and 13 classifiers with parameters set as specified in the authors' paper so as to allow comparison: MLP with 3 hidden layers of sizes $[300, 100, 50]$ respectively; Random Forest with $50$ estimators or trees; 5-fold Grid Search with Random Forest classifier and estimator $([50, 60, 70]$; Linear Support Vector Classifier with L1 and L2 penalty and tolerance of $10^{-3}$; Regularized linear classifier with Stochastic Gradient Descent (SGD) learning with regularization term $alpha=10^{-4}$ and L1, L2 and Elastic-Net penalty; Nearest Centroid with Euclidian metric, where classification is done by representing each class with a centroid; Bernoulli Naive Bayes with smoothing parameter $alpha=10^{-2}$; K-means clustering with 2 clusters and L2 penalty; and Logistic Regression classifier with L2 penalty, tolerance of $10^{-4}$ and regularization term of $1.0$. Most often, the best performing classifier was MLP. 

\paragraph{BERT} Conventional BERT is a BERT-Base Uncased model, meaning that it has 12 transformer blocks $L$, hidden size $H$ of 768, and 12 self-attention heads $A$. The model is fine-tuned with our dataset on 2 Titan X GPUs for 3 epochs with Adam Optimizer, learning rate of $2*10^{-5}$, maximum sequence length of $128$, and warm up proportion of $0.1$. The train batch size is 4 for the Twitter Sentiment Corpus and 8 for the Chatbot Intent Classification Corpus. 

\paragraph{Stacked DeBERT} Our proposed model is trained in an end-to-end manner on 2 Titan X GPUs with the same hyperparameters as BERT and training time depending on the size of the dataset and train batch size. The stack of multilayer perceptrons, in the denoising block of our model, are trained for 100 and 1,000 epochs with Adam Optimizer, weight decay of $10^{-5}$, MSE loss criterion and batch size the same as BERT (4 for the Twitter Sentiment Corpus and 8 for the Chatbot Intent Classification Corpus). In this step, we increase the learning rate to $10^{-3}$ for faster training.

\subsection{Results on Sentiment Classification from Incorrect Text} \label{sec:results_twitter}
Experimental results for the Twitter Sentiment Classification task on Kaggle's Sentiment140 Corpus dataset, displayed in Table \ref{tbl:twitter_f1_scores}, show that our model has better F1-micros scores, outperforming the baseline models by 6$\%$ to 8$\%$. We evaluate our model and baseline models on three versions of the dataset. The first one (\textit{Inc}) only considers the original data, containing naturally incorrect tweets, and achieves accuracy of 80$\%$ against BERT's 72$\%$. The second version (\textit{Corr}) considers the corrected tweets, and shows higher accuracy given that it is less noisy. In that version, Stacked DeBERT achieves 82$\%$ accuracy against BERT's 76$\%$, an improvement of 6$\%$. In the last case (\textit{Inc+Corr}), we consider both incorrect and correct tweets as input to the models in hopes of improving performance. However, the accuracy was similar to the first aforementioned version, 80$\%$ for our model and 74$\%$ for the second highest performing model. Since the first and last corpus gave similar performances with our model, we conclude that the Twitter dataset does not require complete sentences to be given as training input, in addition to the original naturally incorrect tweets, in order to better model the noisy sentences.

\begin{table}[ht!]
  \caption{F1-micro scores for the Twitter Sentiment Classification task on Kaggle's Sentiment140 Corpus. Note that: (\textit{Inc}) is the original dataset, with naturally incorrect tweets, (\textit{Corr}) is the corrected version of the dataset and (\textit{Inc+Corr}) contains both. The noise level is represented by the iBLEU score.}
  \label{tbl:twitter_f1_scores}
  \bigskip
  \centering
  \scalebox{0.8}{
  \begin{tabular}{lccc}
    \toprule
     & \multicolumn{3}{c}{\textbf{F1-score} (micro, $\%$)} \\
    \cmidrule(r){2-4}
    \textbf{Model} & \textit{Inc} & \textit{Corr} & \textit{Inc+Corr} \\
    \midrule
      \textit{iBLEU score} & 0.63 & 0.00 & 0.63 \\
    \midrule
    Rasa (spacy) & 44.00 & 54.00 & 54.00 \\
    Rasa (tensorflow) & 53.06 & 60.00 & 59.18 \\
    Dialogflow & 30.00 & 40.00 & 42.00 \\
    SAP Conversational AI & 59.18 & 65.31 & 59.18 \\
    \midrule
    Semantic Hashing & 72.00 & 70.00 & 72.00 \\
    BERT & 72.00 & 76.00 & 74.00 \\
    \midrule
    Stacked DeBERT (ours) & \textbf{80.00} & \textbf{82.00} & \textbf{80.00} \\
    \bottomrule
  \end{tabular}}
\end{table}

\begin{figure}[ht!]
   \centering
       \includegraphics[width=0.75\linewidth]{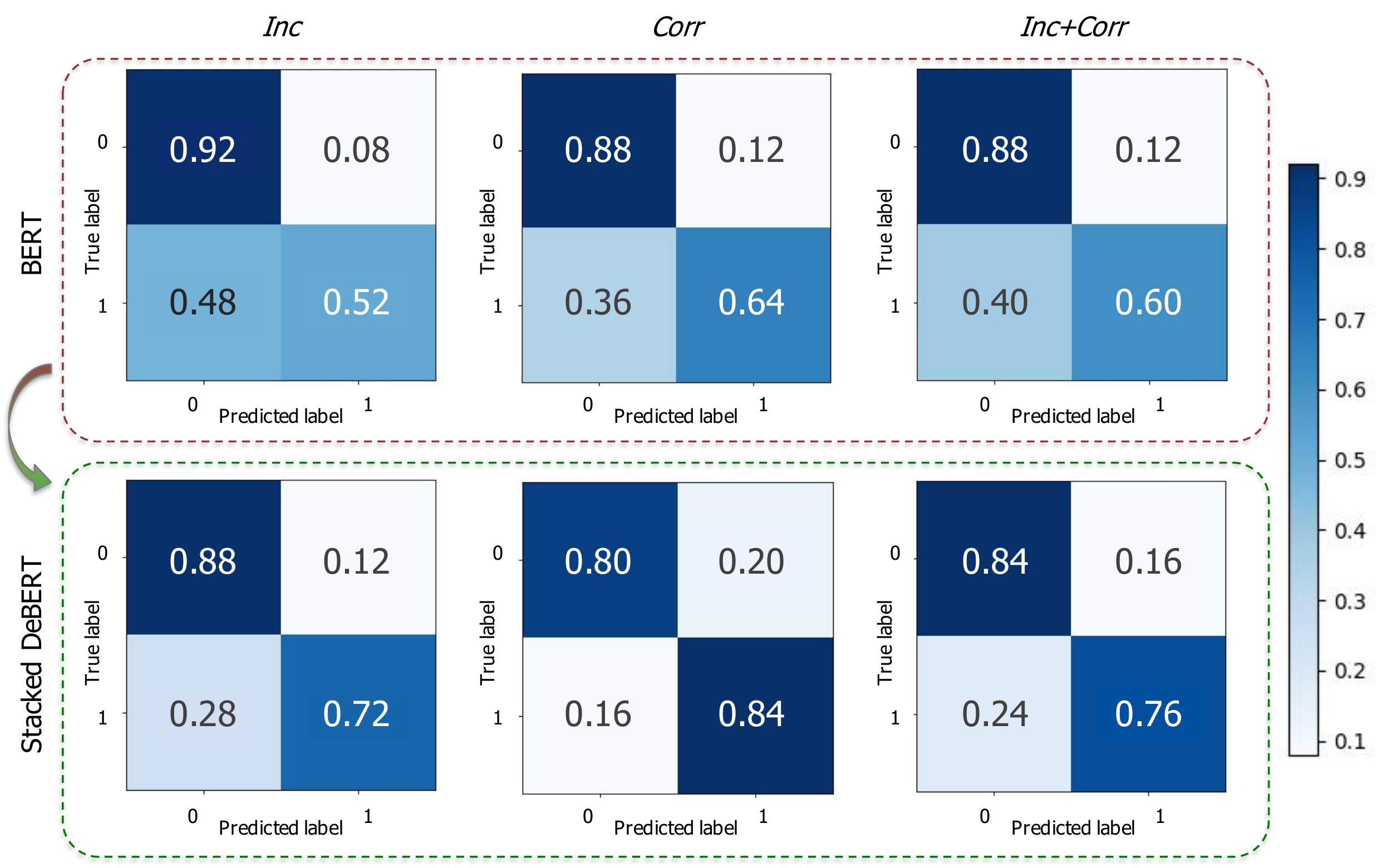}
   \caption{Normalized confusion matrix for the Twitter Sentiment Classification dataset. The first row has the confusion matrices for BERT in the original Twitter dataset (\textit{Inc}), the corrected version (\textit{Corr}) and both original and corrected tweets (\textit{Inc+Corr}) respectively. The second row contains the confusion matrices for Stacked DeBERT in the same order.}
   \label{fig:twitter_confusion_matrix_plot}
\end{figure}

In addition to the overall F1-score, we also present a confusion matrix, in Fig. \ref{fig:twitter_confusion_matrix_plot}, with the per-class F1-scores for BERT and Stacked DeBERT. The normalized confusion matrix plots the predicted labels versus the target/target labels. Similarly to Table \ref{tbl:twitter_f1_scores}, we evaluate our model with the original Twitter dataset, the corrected version and both original and corrected tweets. It can be seen that our model is able to improve the overall performance by improving the accuracy of the lower performing classes. In the \textit{Inc} dataset, the true class 1 in BERT performs with approximately 50\%. However, Stacked DeBERT is able to improve that to 72\%, although to a cost of a small decrease in performance of class 0. A similar situation happens in the remaining two datasets, with improved accuracy in class 0 from 64\% to 84\% and 60\% to 76\% respectively.

\subsection{Results on Intent Classification from Text with STT Error} \label{sec:results_chatbot}

Experimental results for the Intent Classification task on the Chatbot NLU Corpus with STT error can be seen in Table \ref{tbl:chatbot_missing_words_f1}. When presented with data containing STT error, our model outperforms all baseline models in both combinations of TTS-STT: \textit{gtts-witai} outperforms the second placing baseline model by 0.94\% with F1-score of 97.17\%, and \textit{macsay-witai} outperforms the next highest achieving model by 1.89\% with F1-score of 96.23\%.

  \begin{table}[ht!]
   \caption{F1-micro scores for the Chatbot Intent Classification Corpus. Note that we include results with the original sentences (complete data) and sentences imbued with TTS-STT error (\textit{gtts-witai} and \textit{macsay-witai}), with the noise level being represented by the iBLEU and WER scores.}
   \label{tbl:chatbot_missing_words_f1}
   \bigskip
   \centering
   \scalebox{0.8}{
   \begin{tabular}{lccc}
      \toprule
       & \multicolumn{3}{c}{\textbf{F1-score} (micro, $\%$)} \\
      \cmidrule(r){2-4}
      \textbf{Model} & Complete & gtts-witai & macsay-witai \\
      \midrule
      \textit{iBLEU score} & 0.00 & 0.44 & 0.50 \\
      \textit{WER score} & 0.00 & 2.39 & 3.11 \\
      \midrule
      Rasa (spacy) & 92.45 & 91.51 & 86.79 \\
      Rasa (tensorflow) & \textbf{99.06} & 92.89 & 91.51 \\
      Dialogflow & 96.23 & 87.74 & 81.13 \\
      SAP Conversational AI & 95.24 & 94.29 & 94.29 \\
      \midrule
      Semantic Hashing & \textbf{99.06} & 95.28 & 91.51 \\
      BERT & 98.11 & 96.23 & 94.34 \\
      \midrule
      Stacked DeBERT (ours) & \textbf{99.06} & \textbf{97.17} & \textbf{96.23} \\
      \bottomrule
   \end{tabular}}
 \end{table}

\begin{figure}[ht!]
   \centering
       \includegraphics[width=0.75\linewidth]{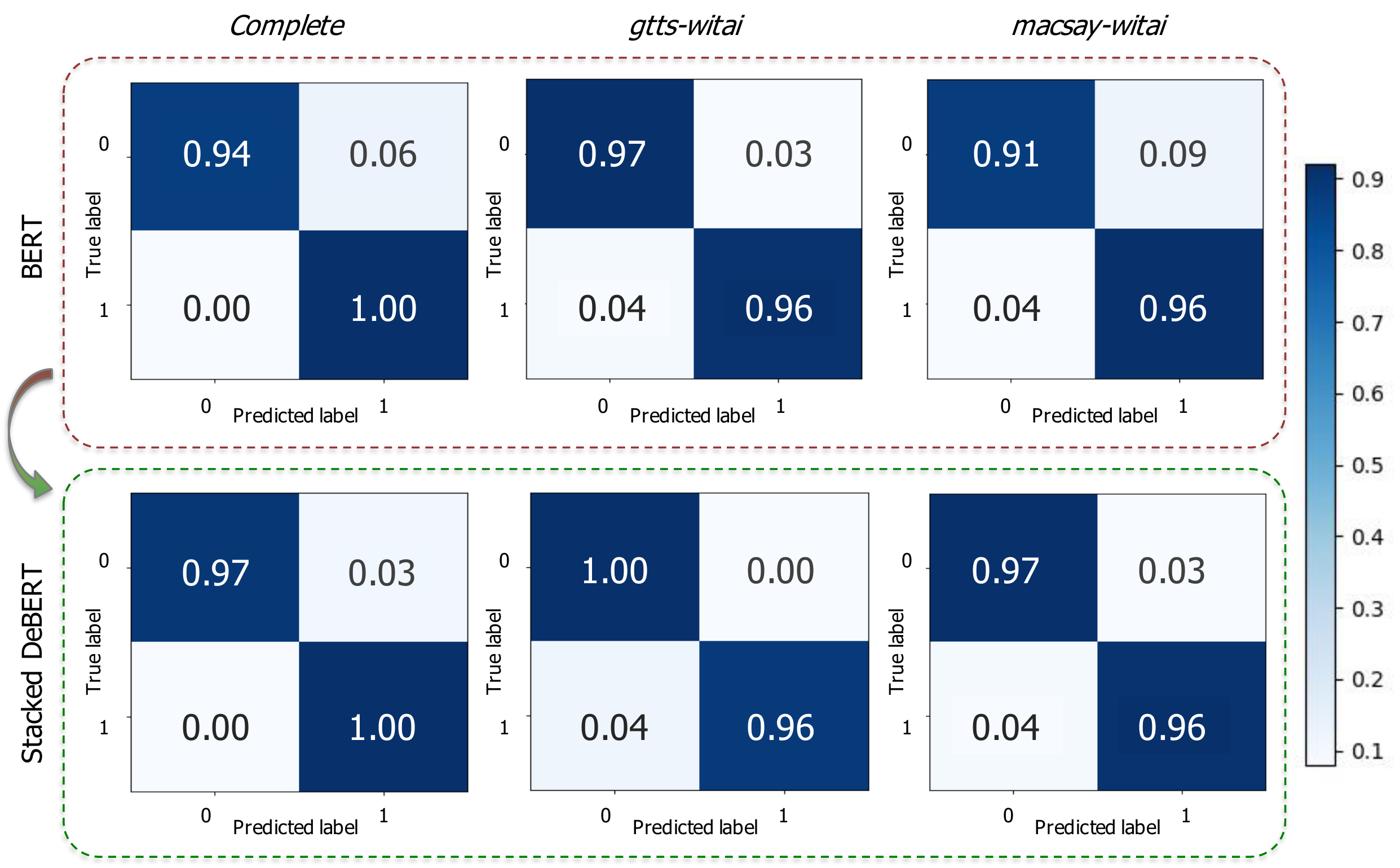}
   \caption{Normalized confusion matrix for the Chatbot NLU Intent Classification dataset for complete data and data with STT error. The first column has the confusion matrices for BERT and the second for Stacked DeBERT.}
   \label{fig:chatbot_stterror_confusion_matrix}
\end{figure}

Additionally, we also present Fig. \ref{fig:chatbot_stterror_confusion_matrix} with the normalized confusion matrices for BERT and Stacked DeBERT for sentences containing STT error. Analogously to the Twitter Sentiment Classification task, the per-class F1-scores show that our model is able to improve the overall performance by improving the accuracy of one class while maintaining the high-achieving accuracy of the second one.

\section{Result Analysis and Discussion}~\label{sec:discussion}
\subsection{Model Robustness}
\begin{figure}[ht!]
   \centering
       \includegraphics[width=0.8\linewidth]{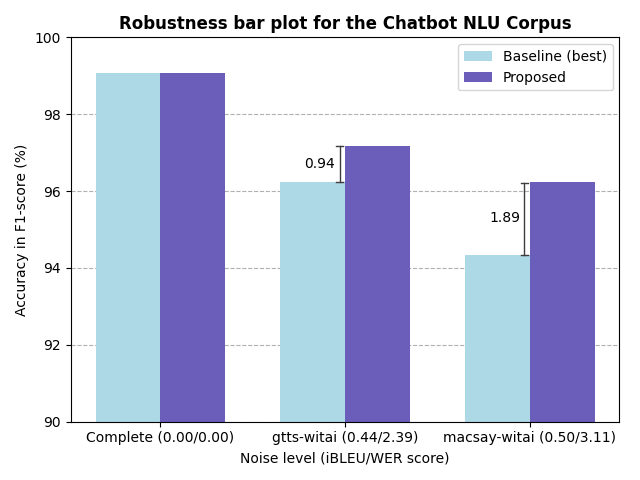}
   \caption{Robustness bar plot for the Chatbot NLU Corpus with STT error.}
   \label{fig:chatbot_stterror_robustness_curve}
\end{figure}

Here we analyze the robustness of our model given varying levels of noise in the Chatbot Intent Classification Corpora. Table~\ref{tbl:chatbot_missing_words_f1} indicates the level of noise in each TTS-STT noisy dataset with their respective iBLEU and WER scores, where 0 means no noise and higher values mean higher quantity of noise. As expected, the models' accuracy degrade with the increase in noise, thus F1-scores of \textit{gtts-witai} are higher than \textit{macsay-witai}. However, our model does not only outperform the baseline models but does so with a wider margin. This is shown with the increasing robustness plot in Fig.~\ref{fig:chatbot_stterror_robustness_curve} and can be demonstrated by \textit{macsay-witai} outperforming the baseline models by twice the gap achieved by \textit{gtts-witai}.

Further analysis of the results in Table \ref{tbl:chatbot_missing_words_f1} show that, BERT decay is almost constant with the addition of noise, with the difference between the complete data and \textit{gtts-witai} being 1.88 and \textit{gtts-witai} and \textit{macsay-witai} being 1.89. Whereas in Stacked DeBERT, that difference is 1.89 and 0.94 respectively. This is stronger indication of our model's robustness in the presence of noise.

\subsection{Performance Comparison with Macro-Averaged Scores}
In this section, we evaluate the performance of our model in more detail by adding a more extensive investigation of the results which include: clearer explanation of the confusion matrix and macro-average precision, recall, and F1 scores. Previously, in Sections~\ref{sec:results_twitter} and~\ref{sec:results_chatbot}, we only included the micro-averaged F1 score since it is considered a good measure of overall effectiveness of classifiers, and it is thus the conventional evaluation metric~\cite{sokolova2009systematic}.

\begin{figure}
    \centering
    \includegraphics[width=0.5\textwidth]{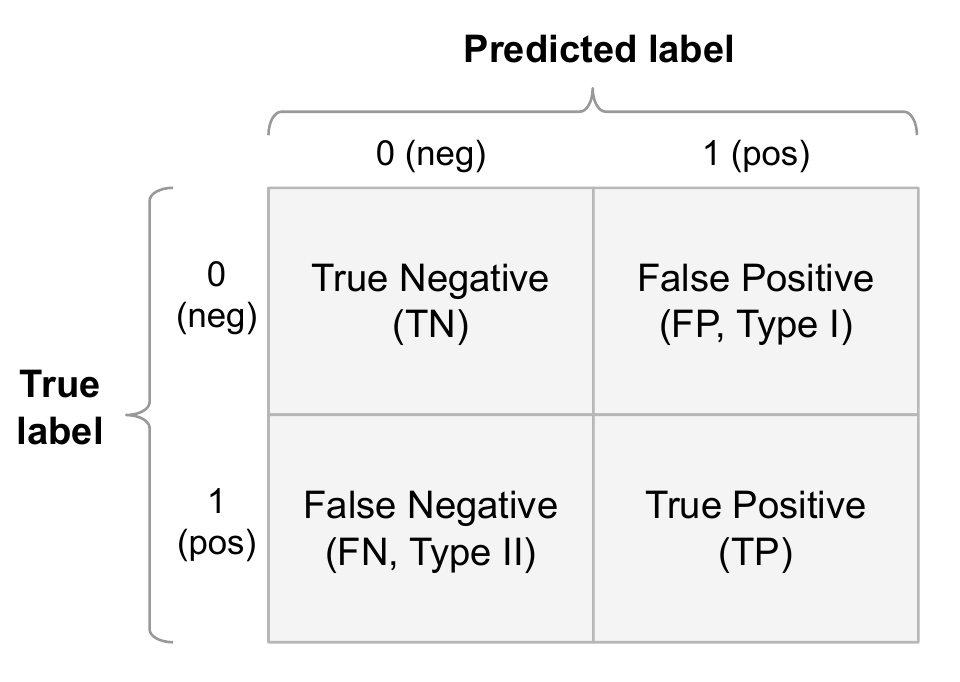}
    \caption{Confusion matrix format for the sentiment classification task, with rows indicating true labels and columns indicating predicted labels, and 0 representing the negative class and 1 representing the positive class.}
    \label{fig:confusion_matrix_explanation}
\end{figure}

To ensure that our following explanation is clear, please consider the confusion matrix~\cite{sokolova2009systematic}, in Fig.~\ref{fig:confusion_matrix_explanation}, for our sentiment classification task. Since our problem is a multi-class problem, where each data sample is assigned to exactly one class, the micro-averaged measures are as in Eq.~\eqref{eq:micro_f1}:
\begin{equation}~\label{eq:micro_f1}
F1 = P = R = \frac{TP+TN}{TP+TN+FP+FN}
\end{equation}
where $P$ is precision, $R$ is recall, and $TP$, $TN$, $FP$, and $FN$ are as indicated in Fig~\ref{fig:confusion_matrix_explanation}. Since the micro-averaged scores are the same, we simply display the micro-F1 score as per usual in the literature. We also further analyze the confusion matrices obtained from both datasets in regards to Type I and II errors. In the Twitter sentiment classification task (Fig.~\ref{tbl:twitter_f1_scores}), our model achieves better overall performance by trading-off accuracy with the best performing class 0, meaning that our model performs slightly worse in the TN and Type I error and significantly better in the Type II and TP errors when compared to the baseline model. In the Chatbot intent classification task (Fig.~\ref{tbl:chatbot_missing_words_f1}), our model performs better or similarly in all instances.

When considering the macro average, however, we are able to separate the influence of each wrong and correct prediction into the precision, recall, and F1 metrics. This allows for a more rigorous performance investigation of our model and the baseline model BERT. Once again, consider the twitter sentiment classification task, where our goal is to classify an input data into positive (1) or negative (0) class. The macro-average measures are the average between all per-class measures. The macro-precision $P_{macro}$ indicates what proportion of positive (or negative) predictions was actually correct. In other words, how precise or accurate the model is. Mathematically, we have Eq.~\eqref{eq:macro_precision}:
\begin{equation} \label{eq:macro_precision}
P_1 = \frac{TP}{TP + FP}, \qquad P_0 = \frac{TN}{TN + FN}, \qquad P_{macro} = \frac{P_1 + P_0}{2}
\end{equation}

The macro-recall $R_{macro}$ indicates what proportion of true positives (or true negatives) was correctly predicted. In other words, how reliable the model is in its predictions. Mathematically, we have Eq.~\eqref{eq:macro_recall}:
\begin{equation} \label{eq:macro_recall}
R_1 = \frac{TP}{TP + FN}, \qquad R_0 = \frac{TN}{TN + FP}, \qquad R_{macro} = \frac{R_1 + R_0}{2}
\end{equation}

Finally, the macro-F1 score combines precision and recall by calculating their harmonic mean and it is thus the preferred metric to indicate the overall performance of a model. Mathematically, we have Eq.~\eqref{eq:macro_f1}:
\begin{equation} \label{eq:macro_f1}
F1_{macro} = 2 * \frac{P_{macro} * R_{macro}}{P_{macro} + R_{macro}} 
\end{equation}

We calculate the macro-average precision, recall, and F1 scores with each dataset and show that our model outperforms the baseline in all cases. Note that, following the same pattern as the micro-F1 scores, our model significantly outperforms the baseline model in the Twitter sentiment classification task and it outperforms the baseline model in the Chatbot intent classification task with a smaller margin. These results are shown in Tables~\ref{tbl:macro_results_twitter} and~\ref{tbl:macro_results_chatbot}.

\begin{table}[ht!]
    \centering
    \caption{Macro-average precision (P), recall (R), and F1 scores (\%) for the Twitter Sentiment Classification Corpus. Note that: (\textit{Inc}) is the original dataset, with naturally incorrect tweets, (\textit{Corr}) is the corrected version of the dataset and (\textit{Inc+Corr}) contains both.}
    \label{tbl:macro_results_twitter}
    \scalebox{0.7}{
    \begin{tabular}{l ccc ccc ccc}
        \toprule
         & \multicolumn{3}{c}{\textit{Inc}} & \multicolumn{3}{c}{\textit{Corr}} & \multicolumn{3}{c}{\textit{Inc+Corr}} \\
        \cmidrule(r){2-4}\cmidrule(r){5-7}\cmidrule(r){8-10}
         Model & P & R & F1 & P & R & F1 & P & R & F1  \\
        \midrule
         BERT & 76.19 & 72.00 & 70.83 & 77.59 & 76.00 & 75.65 & 76.04 & 74.00 & 73.48 \\
         Stacked DeBERT (ours) & \textbf{80.79} & \textbf{80.00} & \textbf{79.87} & \textbf{82.05} & \textbf{82.00} & \textbf{81.99} & \textbf{80.19} & \textbf{80.00} & \textbf{79.97} \\
        \bottomrule
    \end{tabular}
    }
\end{table}

\begin{table}[ht!]
    \centering
    \caption{Macro-average precision (P), recall (R), and F1 scores (\%) for the Chatbot Intent Classification Corpus with the original sentences (complete data) and sentences imbued with TTS-STT error (\textit{gtts-witai} and \textit{macsay-witai}).}
    \label{tbl:macro_results_chatbot}
    \scalebox{0.7}{
    \begin{tabular}{l ccc ccc ccc}
        \toprule
         & \multicolumn{3}{c}{Complete} & \multicolumn{3}{c}{gtts-witai} & \multicolumn{3}{c}{macsay-witai} \\
        \cmidrule(r){2-4}\cmidrule(r){5-7}\cmidrule(r){8-10}
         Model & P & R & F1 & P & R & F1 & P & R & F1  \\
        \midrule
         BERT & 98.63 & 97.14 & 97.83 & 95.22 & 96.46 & 95.79 & 93.60 & 93.60 & 93.60 \\
         Stacked DeBERT (ours) & \textbf{99.31} & \textbf{98.57} & \textbf{98.93} & \textbf{96.05} & \textbf{97.89} & \textbf{96.87} & \textbf{95.22} & \textbf{96.46} & \textbf{95.79} \\
        \bottomrule
    \end{tabular}
    }
\end{table}

\subsection{Performance Improvement Comparison Between Datasets}
When comparing the improvement in performance in the Twitter and TTS-STT Chatbot datasets, we notice that the former shows major improvements whilst the latter shows only minor improvements. We investigate if this is due to lower noise levels in the Twitter dataset. However, as can be seen in Table~\ref{tbl:noise_level_comparison_performance}, our model’s better performance in the Twitter dataset is not related to it having lower noise levels when compared to the TTS-STT Chatbot corpus. A possible reason as to why our model is able to improve its performance by a larger margin in the Twitter dataset can be due to BERT being trained on the Wikipedia and Book Corpus~\cite{devlin2018bert}. Twitter has arguably more noise when compared to BERT’s original training data due to its highly informal setting and character limitation. However, studies suggest that social media text has relatively small grammatical disparity when compared to edited text such as Wikipedia~\cite{baldwin2013noisy}, and since they both contain user generated texts, their basic sentence structure is still much more similar than when compared to sentences with STT error. This different word composition in sentences affected with STT error, makes it harder for the model to perform as well as its counterpart trained with user-generated text.

\begin{table}[ht!]
   \centering
   \caption{Comparison of performance improvement in relation to varying levels of noise between the TTS-STT Chatbot datasets, namely \textit{gtts-witai} and \textit{macsay-witai}, and the Twitter Sentiment Dataset with incorrect text (\textit{Inc}). Note that, for fair comparison, the WER score for the Twitter dataset has also been included here, even though that score is usually only used to measure levels of noise in text with STT error, which is only present in the Chatbot corpus.} \label{tbl:noise_level_comparison_performance}
   \scalebox{0.9}{
    \begin{tabular}{cccc}
        \toprule
         & \multicolumn{2}{c}{\textbf{Chatbot}} & \textbf{Twitter} \\ \cmidrule(lr){2-3}\cmidrule(lr){4-4}
         & gtts-witai & macsay-witai & \textit{Inc}\\
         \hline
         iBLEU & 0.44 & 0.50 & 0.63\\
         WER & 2.39 & 3.11 & 6.36 \\
         \hline
         Improvement & +0.94 & +1.89 & +8.00 \\
        \bottomrule
    \end{tabular}
    }
\end{table}

Another interesting observation in the Twitter dataset is that our proposed model more significantly improves the performance of class 1 (positive) with a small decrease of performance of class 0 (negative). However, the same pattern is not observed when we compare performance in the Chatbot dataset, where the proposed model shows minor improvements in class 0 (Departure Time intent) while maintaining the performance of class 1 (Find Connection intent). We believe that the in-class improvement of class 1 noticed in the Twitter dataset is due to the existence of a larger gap in performance between classes 0 and 1 in the baseline BERT model, which allows for our model to achieve better overall accuracy with a trade-off of performances. After an investigation of possible reasons why that gap in performance exists in the Twitter dataset, we conclude that that is due to differing average number of words in sentences from positive/negative classes and from train/test sets. Table~\ref{tbl:average_number_of_words_per_sentence} shows a comparison between the average number of words in the Twitter dataset and both TTS-STT Chatbot copora. As can be seen, the Chatbot corpus consistently maintains an average number of 9 to 10 words per sentence for both classes in both the train and test sets. Whereas the Twitter dataset has an average of 16 words for both classes during training and an average of 24 words for the negative class versus 9 words for the positive during testing, which explains the better performance in the positive class. A larger test set might also be influencing the smaller performance improvement detected in the Chatbot corpus (see Tables~\ref{tbl:corpus_info_twitter} and~\ref{tbl:corpus_info}).

\begin{table}[ht!]
   \centering
   \caption{Average number of words per sentence in the TTS-STT Chatbot datasets, namely \textit{gtts-witai} and \textit{macsay-witai}, and the Twitter Sentiment Dataset with incorrect text (\textit{Inc}).} \label{tbl:average_number_of_words_per_sentence}
   \scalebox{0.9}{
    \begin{tabular}{cccccccc}
        \toprule
         & & \multicolumn{4}{c}{\textbf{Chatbot}} & \multicolumn{2}{c}{\textbf{Twitter}} \\ \cmidrule(lr){3-6}\cmidrule(lr){7-8}
         & & \multicolumn{2}{c}{gtts-witai} & \multicolumn{2}{c}{macsay-witai} & \multicolumn{2}{c}{\textit{Inc}} \\ 
         & \textbf{Class} & Train & Test & Train & Test & Train & Test \\
         \hline
         \textbf{Avg. \#} & 0 & 9 & 9 & 9 & 9 & 16 & 24 \\
         \textbf{words} & 1 & 9 & 10 & 10 & 10 & 16 & 9 \\
        \bottomrule
    \end{tabular}
    }
\end{table}

\section{Conclusion}
\label{sec:conclusion}
In this work, we proposed a novel deep neural network, robust to noisy text in the form of sentences with missing and/or incorrect words, called Stacked DeBERT. The idea was to improve the accuracy performance by improving the representation ability of the model with the implementation of novel denoising transformers. More specifically, our model was able to reconstruct hidden embeddings from their respective incomplete hidden embeddings. Stacked DeBERT was compared against three NLU service platforms and two other machine learning methods, namely BERT and Semantic Hashing with neural classifier. Our model showed better performance when evaluated on F1 scores in both Twitter sentiment and intent text with STT error classification tasks. The per-class F1 score was also evaluated in the form of normalized confusion matrices, showing that our model was able to improve the overall performance by better balancing the accuracy of each class, trading-off small decreases in high achieving class for significant improvements in lower performing ones. In the Chatbot dataset, accuracy improvement was achieved even without trade-off, with the highest achieving classes maintaining their accuracy while the lower achieving class saw improvement. Further evaluation on the F1-scores decay in the presence of noise demonstrated that our model is more robust than the baseline models when considering noisy data, be that in the form of incorrect sentences or sentences with STT error. Not only that, experiments on the Twitter dataset also showed improved accuracy in clean data, with complete sentences. We infer that this is due to our model being able to extract richer data representations from the input data regardless of the completeness of the sentence. For future works, we plan on evaluating the robustness of our model against other types of noise, such as word reordering, word insertion, and spelling mistakes in sentences. In order to improve the performance of our model, further experiments will be done in search for more appropriate hyperparameters and more complex neural classifiers to substitute the last feedforward network layer.

\section*{Acknowledgments}
This work was partly supported by Institute of Information \& Communications Technology Planning \& Evaluation (IITP) grant funded by the Korea government (MSIT) (2016-0-00564, Development of Intelligent Interaction Technology Based on Context Awareness and Human Intention Understanding) and Korea Evaluation Institute of Industrial Technology (KEIT) grant funded by the Korea government (MOTIE) (50\%) and the Technology Innovation Program: Industrial Strategic Technology Development Program (No: 10073162) funded By the Ministry of Trade, Industry \& Energy (MOTIE, Korea) (50\%).

\bibliographystyle{unsrt}
\bibliography{mybibfile.bib}


\end{document}